\documentclass[letterpaper, 10 pt, conference]{ieeeconf}  

\usepackage[T1]{fontenc} 

\usepackage{graphicx}
\graphicspath{ {./figures/} }

\usepackage{hyperref}
\usepackage[multiple]{footmisc} 
\usepackage{caption}
\usepackage{subfig} 
\usepackage{mathtools}
\usepackage{gensymb} 
\usepackage{cite}
\usepackage{amssymb}
\usepackage{algorithm} 
\usepackage{algpseudocode} 
\usepackage{booktabs}

\IEEEoverridecommandlockouts                              

\title{\LARGE \bf
I Know You Can’t See Me: Dynamic Occlusion-Aware Safety Validation of Strategic Planners for Autonomous Vehicles Using Hypergames
}

\author{Maximilian Kahn,  Atrisha Sarkar, and Krzysztof Czarnecki
\thanks{M. Kahn, A. Sarkar, and K. Czarnecki are with the University of Waterloo, 200 University Ave W, Waterloo, ON, N2L3G1 \newline
        {\tt\small \big\{maximilian.kahn, atrisha.sarkar, krzysztof.czarnecki\big\}@uwaterloo.ca}.}
}

\begin{document}

\maketitle
\thispagestyle{empty}
\pagestyle{empty}

\begin{abstract}
A particular challenge for both autonomous and human driving is dealing with risk associated with dynamic occlusion, i.e., occlusion caused by other vehicles in traffic. Based on the theory of hypergames, we develop a novel multi-agent dynamic occlusion risk (DOR) measure for assessing situational risk in dynamic occlusion scenarios. Furthermore, we present a white-box, scenario-based, accelerated safety validation framework for assessing safety of strategic planners in AV. Based on evaluation over a large naturalistic database, our proposed validation method achieves a $4000$\% speedup compared to direct validation on naturalistic data, a more diverse coverage, and ability to generalize beyond the dataset and generate commonly observed dynamic occlusion crashes in traffic in an automated manner.
\end{abstract}

\section{INTRODUCTION}
Safety validation of autonomous vehicle (AV) planners are a critical component in the development of AVs. In recent years, as AVs face the challenge of sharing the roads with other human drivers with diverse behavior, the problem of behavior planning for AVs has taken a multi-agent view with the use of game-theoretic models for planning \cite{li2018game,tian2018adaptive,tian2021anytime,li2019decision,Geiger_Straehle_2021,fisac2019hierarchical}. Such planners, referred to as \emph{strategic} planners, view other road users in the vicinity as agents playing a game, and the AV chooses an action based on a game solution, for example, a Nash equilibrium. Such models have shown to be effective in simulation, and also have been evaluated against naturalistic human driving behavior \cite{sarkar2021solution, sun2020game}. On the other hand, evaluation of safety is also arguably a multi-agent problem, based on the idea that the outcome of a traffic situation depends on the assumptions traffic agents have over each other as well as the collective behavior based on those assumptions. For example, it is clear that two vehicles deciding to cross an intersection at the same time results in a higher risk than one vehicle deciding to wait for the other. In order to perform a white-box safety assessment of strategic planners, a safety validation framework needs to be aware that the AV planner places certain assumptions on the model of behavior of other drivers, and therefore ideally focus the safety assessment on scenarios where such assumptions are likely to break down. Since most existing game theoretic planners work under the assumption that each vehicle in the game is aware of all other vehicles in traffic, occlusions are prime avenues of such high risk situations. \par
Occlusions or obstructed views, where a road user’s view is obstructed due to static structures (trees, buildings, etc.), other vehicles in traffic, and road elevation and geometry, are a major cause of traffic accidents \cite{national2008national}. The majority of the assessment of risk from occlusion has been focused on \emph{static} occlusion, i.e., occlusions that are caused by static structures, such as trees, buildings, parked cars, etc. \cite{yu2019occlusion, mcgill2019probabilistic, damerow2017risk}. On the other hand, situations of \emph{dynamic} occlusion, i.e., occlusion caused by another vehicle in traffic, have unique challenges and can appear unexpectedly at any moment in traffic. For the problem of planning under dynamic occlusions, existing methods have been proposed with a single-agent view (i.e., without taking into account the collective behavior) \cite{hubmann2019pomdp,lin2019decision,bouton2018scalable};
however, to our knowledge, there are no safety validation frameworks for evaluation of strategic planners for the problem. Given the criticality of dynamic occlusion scenarios \cite{national2008national, choi2010crash}, there is a need to address this gap. \par
A major challenge of dealing with dynamic occlusions is that they are transient in nature, i.e., they can occur at any traffic situation based on certain alignments of three vehicles. Therefore, unlike the case of static occlusions, it is not possible to leverage prior information about structures, such as buildings and trees, from high definition (HD) maps and incorporate that information in the assessment of occlusion risk. The transient nature also means that the scope of the problem is much bigger, with infinite possibilities of dynamic occlusions causing potentially risky situations that can arise in traffic. Therefore, existing repositories of observational data drawn from naturalistic driving are not adequate to achieve complete coverage of all dynamic occlusion situations an AV will face in its operational lifetime. To address the problem at scale, following the principle of a scenario-based accelerated evaluation \cite{zhao2016accelerated,riedmaier2020survey}, an ideal approach should generate realistic as well as critical test scenarios of simulations that can effectively assess the safety based on situations that goes beyond the set of observed scenarios in existing data. \par
In this paper, we address the aforementioned challenges by making the following contributions. i) A novel \emph{planner-in-the-loop} (white-box) safety validation framework for strategic planners using the theory of \emph{hypergames} ii) A multi-agent dynamic occlusion risk ($DOR$) measure for assessing situational risk in dynamic occlusion scenarios, and iii) A search-based method of augmenting naturalistic data with realistic dynamic occlusion scenarios using vehicle injection.\par
While dynamic occlusion can occur anywhere while driving, occlusions tend to occur at multi-lane intersections where there is ample opportunity for vehicles to block each other's view and vehicles are often moving along paths that intersect. In fact, the National Motor Vehicle Crash Causation Survey found that 7.8\% of all intersection-related collisions were caused by the driver's incorrect decision to turn with an obstructed view of traffic \cite{choi2010crash}. Therefore, we demonstrate the efficacy of the approach with the help of experiments conducted on a large naturalistic dataset from a busy traffic intersection, and show that our proposed validation method achieves a $4000$\% gain in generating occlusion causing crashes compared to naturalistic data only, along with a more diverse coverage, and ability to generate commonly observed dynamic occlusion crashes in traffic that are beyond the naturalistic data used as input in an automated manner.\par
\section{BACKGROUND}
Strategic planners for AV planning use standard models of game theory with the assumption that all agents have a common view of the game, including the set of available actions to each player, the utility of each action for every agent, collectively are a part of \emph{common-knowledge} in game-theoretic terminology \cite{geanakoplos1992common}. However, due to factors such as occlusion, distraction, and inattention, a driver may not be aware of the presence of another conflicting vehicle, and thereby have a different view of the game than other vehicles in the vicinity. The framework of hypergames provides a formal model of interaction for such scenarios where the strict assumption of a common view of the game breaks down \cite{bennett1977toward, kovach2015hypergame, wang1988modeling}. \par
In a standard formulation of a game, all agents play a common game $G = (N,A,U)$, where $N$ is the number of agents in the game indexed by $i$, $A = \prod\limits_{\forall i \in N}A_{i}$ is the set of actions available to all agents, and $U : A \rightarrow R^{N}$ are the utilities that map a set of actions of every agent (a strategy) to a real vector $R^N$, and $U_{i}$ is the $i^{\text{th}}$ component of the vector representing the utility of the strategy to player $i$. In the hypergame framework, instead of agents playing a common game $G$, they play \emph{hypergames} ($H$). The hypergames $H = \{H^{0}, H^{1}, ..,H^{L}\}$ are organized in levels of hierarchy of games, where at higher levels, agents have higher awareness about other agents’ view of the game that may not match their own. The level-0 hypergame is the common singular game $G$, i.e., $H^{0} = G$, where all agents share the common view and play the common game. At level 1, players have differing views of the game, i.e., $H^{1} = \{G^{1}_{1},G^{1}_{2},..,G^{1}_{N}\}$, where $G^{1}_{i} = (N^{1},A^{1},U^{1})_{i}$ is the $i^{\text{th}}$ agent's view of the game, where they may have a completely different view of the number of agents, the actions, and utilities of the game relative to other agents' view $G^{1}_{-i}$ (where $-i$ represents any other agent). At level 2, agents not only have their own individual view of the game, but also awareness that other agents may have their own different views; therefore, the level-2 hypergame   $H^{2} = \{H^{1}_{1},H^{1}_{2},..,H^{1}_{N}\}$, where each agent $i$'s view of the level-1 hypergame, $H^{1}_{i}$, together forms $H^{2}$. Continuing up the hierarchy, the model of hypergames can be extended to a finite level $L$; however, we focus up to level-2 in our analysis, since that level is sufficient to model risk arising out of conflicts due to dynamic occlusion scenarios in traffic.\par
In this paper, we use the above theory to construct three distinct perspectives of a traffic situation with respect to dynamic occlusion: level-0 is \emph{occlusion resolved}, where all agents have a common omniscient view of the traffic situation (i.e., as if vehicles were transparent objects), level-1 is \emph{occlusion naive}, i.e., drivers ignore occluded spaces, and level-2 is \emph{occlusion aware}, i.e., drivers are aware of the occlusions in the situation and can have their own subjective process of resolving these occlusions and incorporating that awareness in their planning. We also assume the simple and challenging setting where only one vehicle in the scene is occlusion-aware, such as an AV under test, and the rest are occlusion-naive human drivers. Therefore, $H^{2} =\{H^{1}\}$ is a singleton set where the level-1 hypergame $H^{1}$ is constructed from the sole occlusion-aware vehicle's perspective.\par
A solution concept provides a solution to a game, which results in a strategy-profile, i.e., a set of trajectories that every vehicle executes. The process of calculating the solution and executing the strategies is referred to as playing or solving the game. Some of the solution concepts that have been proposed in the literature for strategic planning in AVs include Nash equilibrium \cite{pruekprasert2019decision,schwarting2019social,Geiger_Straehle_2021,sarkar2021solution}, Stackelberg equilibrium \cite{fisac2019hierarchical}, Qlk model \cite{tian2018adaptive, tian2021anytime, li2018game, sarkar2021solution}, and Pareto optimality \cite{sun2020game}. \par
The games are instantiated as simultaneous move games based on the joint system state $X_{t} = \prod\limits_{\forall i \in N}X_{i,t}$, of $N$ vehicles in traffic at time $t$. $X_{i,t} = [x,y,v_{x},v_{y},\dot{v_x},\dot{v_y},\theta]$ are location co-ordinates ($x,y$) on $R^{2}$, lateral and longitudinal velocity ($v_{x},v_{y}$) in the body frame, acceleration ($\dot{v_x},\dot{v_y}$), and yaw ($\theta$) of a vehicle $i$ at time $t$. We use the notation $\sigma(X_{t},G)$ for a strategy-profile for all players in the game $G$ played at state $X_{t}$, with an added subscript $\sigma_{i}$ to refer to vehicle $i$'s strategy in the strategy-profile. The actions in the game are cubic spline trajectories \cite{kelly2003reactive, al2018spline}, generated over a planning horizon of 6 secs. The utilities $U_{i}$ in the game are multi-objective, with two components---safety (a sigmoidal function that maps the minimum distance gap between vehicle trajectories into the interval [-1,1]), and progress (a linear function that maps the length of the trajectory in meters into the interval [0,1]). The two objectives are combined using a lexicographic thresholding parameter ($\gamma$) \cite{LiChangjian19}. Following \cite{fisac2019hierarchical, sarkar2021solution}, we also use a hierarchical decomposition over the actions of the game where trajectories are generated based on high-level maneuvers. We refer to \cite{sarkar2021solution} for the detailed construction of the game. Since our validation method is white-box, we need to choose a solution concept a strategic planner uses. We choose Nash equilibrium (for the level of maneuvers) due to its ubiquity \cite{pruekprasert2019decision,schwarting2019social,Geiger_Straehle_2021,sarkar2021solution} along with maxmax (for the level of trajectories) for its promise of being able to model naturalistic human driving behavior better than others \cite{sarkar2021solution}. However, we note the method is planner agnostic, and can be extended to any strategic planner with just the knowledge of the solution concept.
\begin{figure}[t]
\centering
\makebox[250pt]{\subfloat[]{\includegraphics[ width=0.5\linewidth]{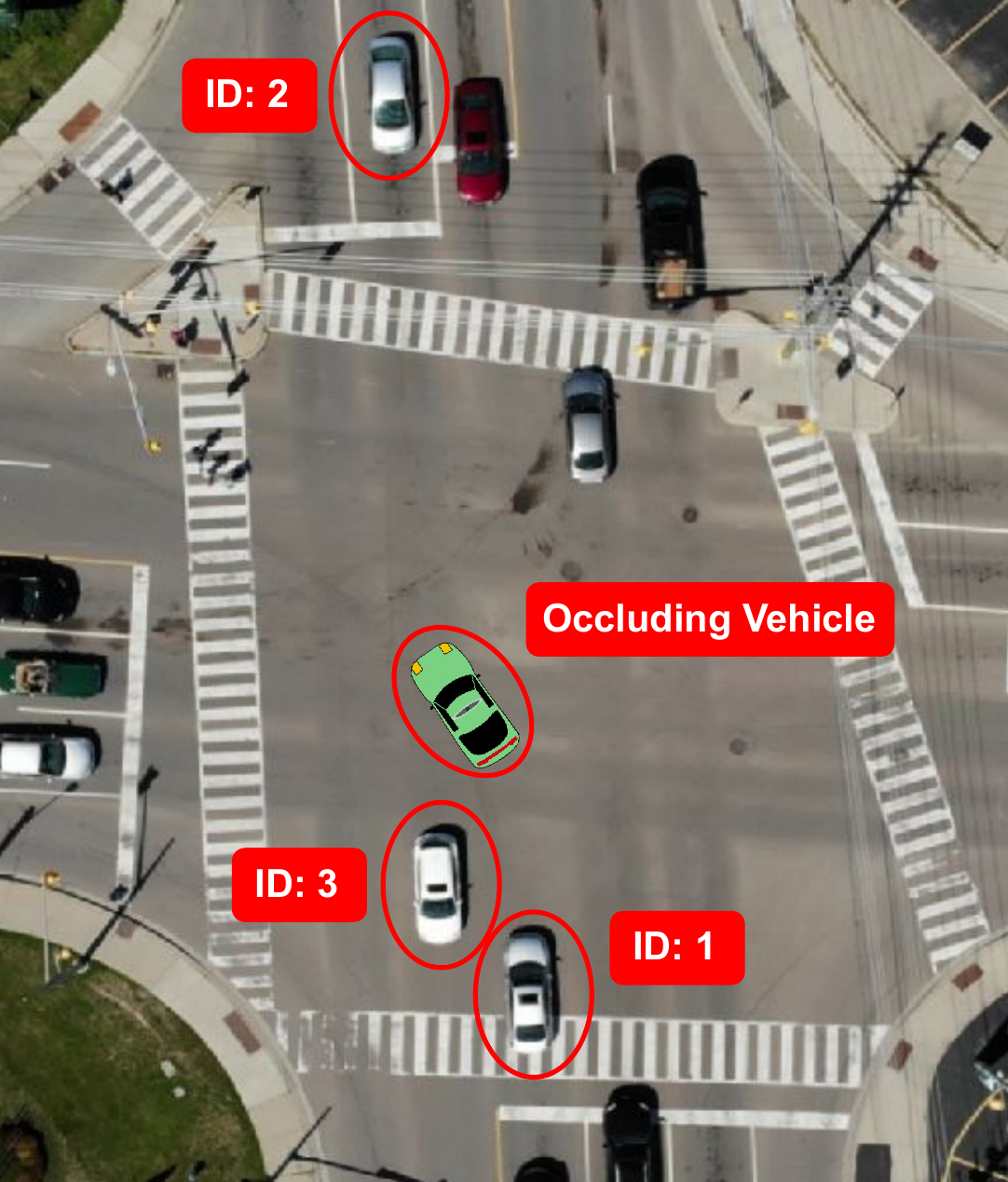}
\label{fig:Full Scenario 778 4243}}}
\newline
\subfloat[]{\includegraphics[ width=0.5\linewidth]{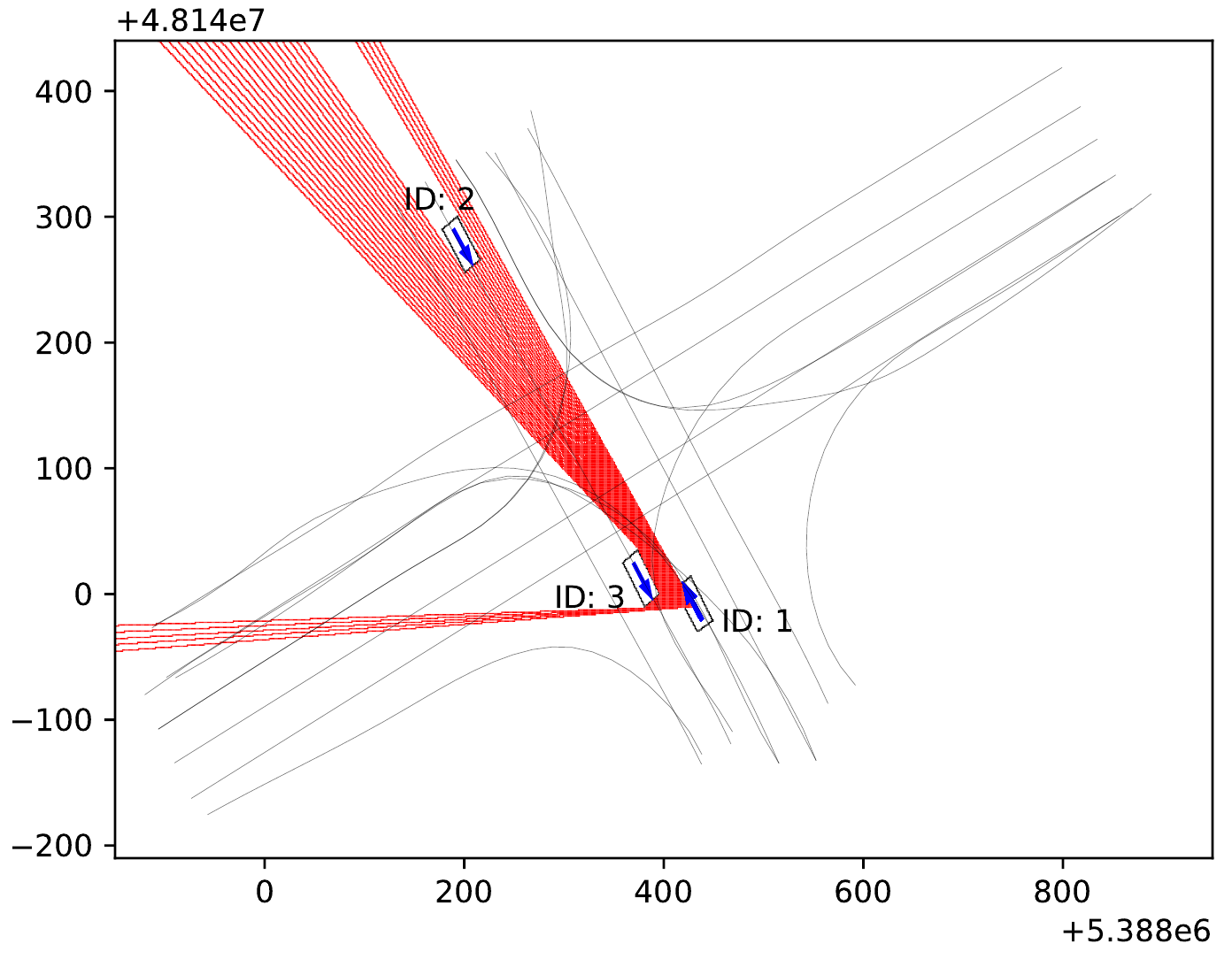}
\label{fig:Partial Scenario With Raycasting}}
\subfloat[]{\includegraphics[ width=0.5\linewidth]{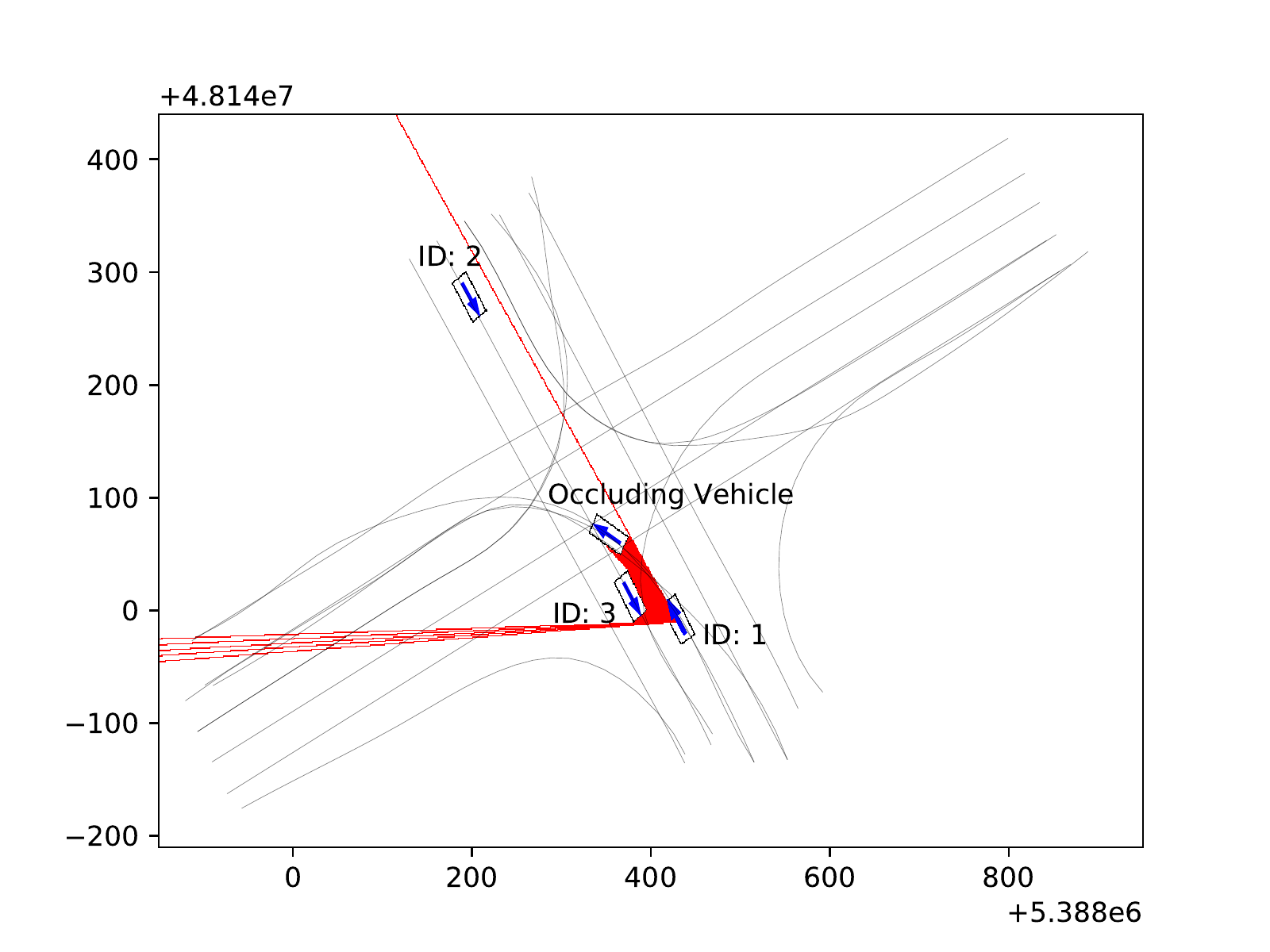}
\label{fig:Occlusion Scenario With Raycasting}}
\captionof{figure}{A left turn across path (LTAP) scenario from the WMA database. (a) real traffic footage along with the injected synthetic OV. (b) occlusion checking without the OV; 2 is unoccluded from 1. (c) occlusion checking with the OV; 2 is occluded from 1. The road lines in (b) and (c) represent the centerlines of each lane.}
\end{figure}
\section{DYNAMIC OCCLUSION}
Occlusions are caused by specific spatial alignments among at least three vehicles, such that one vehicle is obstructed from the view of another vehicle. Let $O(i,j,k) \in \{0,1\}$ be an indicator function representing an occlusion. It has value 1 when vehicle $k$ is occluded from vehicle $i$’s view by an \emph{occluding} vehicle (OV), $j$. For a traffic scenario with $N$ vehicles, if $O(i,j,k) = 0 ;\forall i,j,k \in N$, then the maximum level of the hypergame played by $N$ agents is 0. This follows trivially from the observation that there are no occlusions, and therefore each agent is aware of every other agent in the scene and plays the common game $H^{0} = G$. However, if there is a set of vehicles for which $O(i,j,k) = 1$, there is an occlusion in the scenario, and therefore, on account of $i$ having a different view of the game, the minimum level of hypergame the agents are playing is at least 1. Fig. \ref{fig:Full Scenario 778 4243} illustrates a scenario where $O(1,\text{OV},2) = 1$ and $O(2,\text{OV},1) = 1$, i.e., vehicle 1 and 2 are occluded from each other by vehicle OV.\footnote{The function $O$ is asymmetric since occlusion depends on the position and orientation of the sensors on each vehicle.}\par
Purely the presence of an occlusion does not necessarily lead to a collision; for example, in the same illustrative example, regardless of the view 1 has of the game, if the solution of that game is such that 1 waits for the OV to cross, then that is a much safer outcome compared to a solution where 1 decides to follow the OV across 2’s path and 2 decides to accelerate assuming the OV will cross the intersection in time (recall that 1 and 2 are not aware of each other). Therefore, we develop our risk estimation with a \emph{planner-in-the-loop} approach that takes into account along with the traffic situation, the dynamic behavior of all the involved vehicles given by the strategic model that the AV planner uses. Next, we present this notion more formally. \par
An occlusion-aware perspective is the level-1 hypergame $H^{1} = \{G^{1}_{1},G^{1}_{2},..,G^{1}_{N}\}$ representing the individual games that occlusion-naive vehicles play. A Dynamic Occlusion Risk (DOR) is a measure of relative risk based on that hypergame and the traffic state $X_{t}$ as follows.
\begin{equation}
\small
\begin{aligned}
    DOR(X_{t},H^{1},\mathcal{S}) =& \mathcal{S}(\underbrace{\sigma^{*}(X_{t},H^{0})}_\text{occlusion-resolved strategy}) \\ &- \mathcal{S}(\underbrace{\prod\limits_{\forall i \in N}\sigma_{i}(X_{i,t}\cdot \hat{X}^{i}_{t},G^{1}_{i})}_\text{occlusion naive strategies})
\end{aligned}
\normalsize
\end{equation}
where $\hat{X}^{i}_{t} = \{X_{y,t}: O(i,x,y)=0; \forall x \in N, \forall y \in N \setminus x \}$, or the game state constructed out of the vehicles that an occlusion-naive vehicle can see. The first component of the equation is the safety, with respect to a surrogate metric $\mathcal{S}$ (we use minimum distance gap), of the occlusion-resolved game. This reflects the situation where all vehicles see each other and follow the equilibrium strategy. However, an occlusion-naive vehicle $i$ will solve the game $G^{1}_{i}$ instead, and construct the occlusion naive strategies. Comparing that strategy to the solution of the occlusion-resolved game ($H^{0}$) estimates the relative risk arising from the dynamic occlusion. The above measure is calculated based on one occlusion-aware AV's level-1 perspective ($H^{1}$). However, if there are multiple AVs in the scenario with their own level-1 perspectives, the measure can be calculated from each of those perspectives. The perspectives could be coordinated via V2X and the estimated risk incorporated into planning. 
\section{METHOD}
\begin{figure}[!t]
         \centering
         \includegraphics[trim=0 0 0 -0.8cm, width=\linewidth]{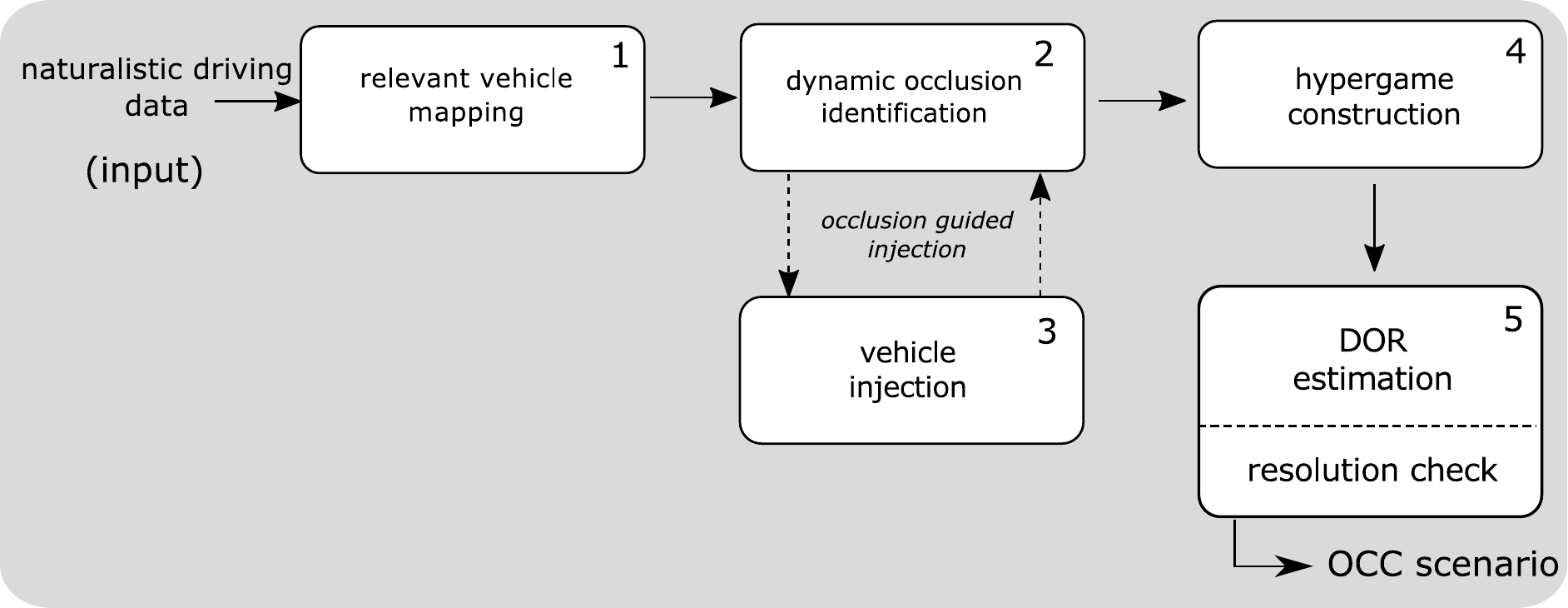}
         \caption{Schematic representation of accelerated scenario-based dynamic occlusion safety validation method}
         \label{fig:method}
\end{figure}
Fig. \ref{fig:method} shows a schematic representation of our proposed method. A scenario-based accelerated evaluation method \cite{zhao2016accelerated, riedmaier2020survey} generally has three key components, all of which are covered in our approach --- a) being able to draw scenarios from naturalistic driving data (input, step 1), b) guided generation or sampling of novel test scenarios based on a target condition (steps 2-3), and c) identification of risk based on a quantifiable measure and assessment of the component under test through the measure (steps 4-5). Next, we describe each of the steps in more detail.
\subsection{Naturalistic data and Relevant vehicle mapping.}
From all the situations in naturalistic driving data (we use the Intersection dataset from the WMA database \cite{sarkar2021solution}), we extract scenarios corresponding to the two main intersection navigation tasks, i.e., LTAP (Fig. \ref{fig:Full Scenario 778 4243}) and unprotected right turn (RT). For each of these scenarios, we select the vehicle executing the scenario at a given time, and call it the \emph{subject} vehicle. With respect to the subject vehicle, we construct the set of \emph{relevant} vehicles by including i) any vehicle that is in cross-path conflict with the subject vehicle, ii) the leading vehicle of the subject vehicle, and iii) the leading vehicle of a cross-path conflicting vehicle. The set of relevant vehicles along with the subject vehicle forms the set of vehicles $N$, the system state $X_t$ at time step $t$, and represents a situation. The set $D$ is the set of all such situations in the dataset and represents important interactions based on traffic conflicts, and we use the set to create the games/hypergames in the subsequent steps. The notion of \emph{subject} and \emph{relevant} vehicle is used here as a way of isolating the LTAP and RT scenarios from the input dataset and do not have any special meaning outside of this context. As a minimal example of this construction, in the snapshot shown in Fig. \ref{fig:Full Scenario 778 4243}, 1 is a subject vehicle corresponding to a LTAP scenario; the relevant vehicles is the set \{2,3\}, since 2 is in cross path conflict with 1, and 3 is 2's leading vehicle.
\subsection{Dynamic occlusion identification and vehicle injection}
\begin{algorithm}[bp]
\caption{Occlusion-guided random search for a naturalistic scene with N vehicles.}
\label{alg:Occlusion-Guided Random Search}
\begin{algorithmic}[1]
\small
\State $D^{\mathcal{O}} \gets \{\emptyset\}$
\For{$s \in D$}:
\State $V^{\mathcal{I}} \gets \text{occluding\_vehicle\_sampling}(s)$
\For{$o \in V^{\mathcal{I}}$}:
    \For{$i \in N$} 
    \If{$\exists x \in N\setminus i; O(i,o,x) = 1$}
        \State $s \gets \text{add\_to\_scene}(o)$
        \State $D^{\mathcal{O}} \gets D^{\mathcal{O}} + \{s\}$
    \EndIf
    \EndFor
\EndFor
\EndFor\newline
\Return $D^{\mathcal{O}}$
\normalsize
\end{algorithmic}
\end{algorithm}

The goal of this stage is twofold; first, to implement an occlusion check process that identifies whether there is a dynamic occlusion in a given traffic scene (step 2), or in other words, implement the occlusion indicator function $O$, and second, augment naturalistic data with such scenes by injecting vehicles in realistic configurations (step 3).\par
To accomplish the first goal, we use a voxel-based raycasting approach \cite{amanatides1987fast}. We plot the raycasts and vehicle bounding boxes (of size $l$ by $w$) using an occupancy grid map with a grid size of $c$. A vehicle is considered occluded if fewer than $\epsilon$ rays collide with its bounding box. A particular challenge of modeling driver vision is modeling driver attention, since \textit{where} a driver is looking directly influences what they can and cannot see. We use a distance-based approach as a model of driver attention, whereby closer vehicles receive more attention than vehicles that are farther away. The intuition behind this is that human drivers pay more attention to vehicles that are nearby since these vehicles are more relevant to the driver's decision-making process. This translates to closer vehicles receiving a larger number of raycasts than vehicles that are farther away. Figs.\,1b,c
are examples of our occlusion-checking approach applied to a LTAP scenario with and without a synthetic occluding vehicle.\par 
Algo. \ref{alg:Occlusion-Guided Random Search} achieves the second goal (step 3). Since dynamic occlusion can occur anywhere in the intersection, we sample configurations of potential occluding vehicles that can be injected realistically in the scene (ln.\,3, \textit{occluding\_vehicle\_sampling}). We use a grid based sampling with a resolution of $d$ meters along the lane centerline, and place vehicles ensuring minimum distance gap with existing vehicles. For velocities, we sample from a distribution based on the naturalistic dataset. Step 3 iterates over the samples of injected vehicles, $V^{\mathcal{I}}$, and if the injected vehicle causes an occlusion, the vehicle's configuration is added to the scene $s$ (ln.\,7, \textit{add\_to\_scene}), and the scene is added to the list of occlusion scenes $D^{\mathcal{O}}$ (ln.\,8).
\begin{algorithm}[bp]
\caption{DOR identification algorithm.}
\label{alg:hyper_constr}
\begin{algorithmic}[1]
\small
\State $D^{\mathcal{C}} \gets \{\emptyset\}$ \Comment{Initialize OCC set}
\For{$s \in D^{\mathcal{O}}$}
    \State $H^{1} \gets \{\emptyset\}$  \Comment{initialize hypergame set} 
    \State $X_{t} \gets \text{states}(s) $ \Comment{construct vehicle states}
    \For{$i \in N$} 
        \State $N_{i} \gets N$
        \For{$j \in N \setminus i$}
            \If{$\exists x: O(i,x,j) = 1$}
                \State $N_{i} \gets N_{i} - \{j\}$ \Comment{Remove occluded vehicle}       
            \EndIf
        \EndFor 
    \State $G^{1}_{i} \gets (N_{i}^{1},A^{1}_{N_{i}},U^{1}_{N_{i}})$ \Comment{occlusion-naive game}
    \State $H^{1} \gets H^{1} + \{G^{1}_{i}\}$ \Comment{Add it to the hypergame}
    \EndFor
    \If{$DOR(X_{t},H^{1},S) \geqslant \theta$} \Comment{Collision check}
        \State $D^{\mathcal{C}} \gets D^{\mathcal{C}} + \{s\}$ \Comment{Add to OCC}
    \EndIf
\EndFor \newline
\Return $D^{\mathcal{C}}$
\normalsize
\end{algorithmic}
\end{algorithm}
\subsection{Hypergame construction and DOR estimation}
Step 5 outputs the set of occlusion-caused collision (OCC) scenarios, represented by their initial scenes $D^{\mathcal{C}} \subseteq D^{\mathcal{O}}$. It does so by selecting the scenes from $D^{\mathcal{O}}$ that lead to collisions based on the DOR measure. For each scene in $D^{\mathcal{O}}$, Algo. \ref{alg:hyper_constr} constructs the level-1 hypergame based on what each occlusion-naive vehicle sees and plays ($G_{i}^{1}$). If the DOR measure from the hypergame results in a collision, the corresponding scene is added to the OCC set $D^{\mathcal{C}}$. \par
In order to make the dynamic occlusion situations more realistic, we finish step 5 with a resolution check (Fig. \ref{fig:method}). We run the game starting from state $X_{t}$ in simulation. If right after instantiation of the game, the occlusion is resolved, and an emergency maneuver by any of the vehicles avoids a crash, we exclude that situation from $D^{\mathcal{C}}$.

\section{EXPERIMENTS AND EVALUATION}
\vspace{1em}
\begin{table}[bp]
\vspace{1.5em}
\begin{tabular}{@{}lll@{}}
\toprule
                                   & Naturalistic data & Our approach \\ \midrule
No. of OCC                         & 2                 & 80           \\
No. of dynamic occlusion scenarios & 1534              & 105,914      \\ \bottomrule
\end{tabular}
\caption{Comparison with validation from only naturalistic data.}
\label{tab:nat_comp}
\end{table}
To evaluate and demonstrate the efficacy of our approach, we use the Waterloo Multi-Agent (WMA) database recorded at a busy Canadian intersection with over 3.5k vehicles. We evaluate our approach by comparing its output to the dynamic occlusion and risk identified directly from naturalistic data, that is, without injecting additional occluders. We also evaluate the realism of the occlusions and crashes identified by our method relative to ones commonly seen in traffic, and provide a severity analysis of the identified crashes. For the experiments, we used the following parameter values: $l =4.1m$, $w=1.8m$, $c=0.1m$, and $\epsilon = 3$.\par
\noindent \textbf{Comparison with naturalistic data.} We first compare the number and variety of occlusion-caused collisions (OCC) that were generated based on our approach with occlusion injection and the OCCs generated purely from the naturalistic dataset. As seen in Table \ref{tab:nat_comp}, using a non-augmented approach purely based on the naturalistic dataset, we can generate only 2 OCCs, whereas using the proposed accelerated approach allows us to increase the total number of OCCs by 4000\%, to a total of 80 OCCs. Similar gains of manifold ($7\times10^{4}$) are also seen in the number of dynamic occlusion scenarios generated by our method.\par
\begin{figure}[htbp]
\centering
\subfloat[]{\includegraphics[width=0.5\linewidth]{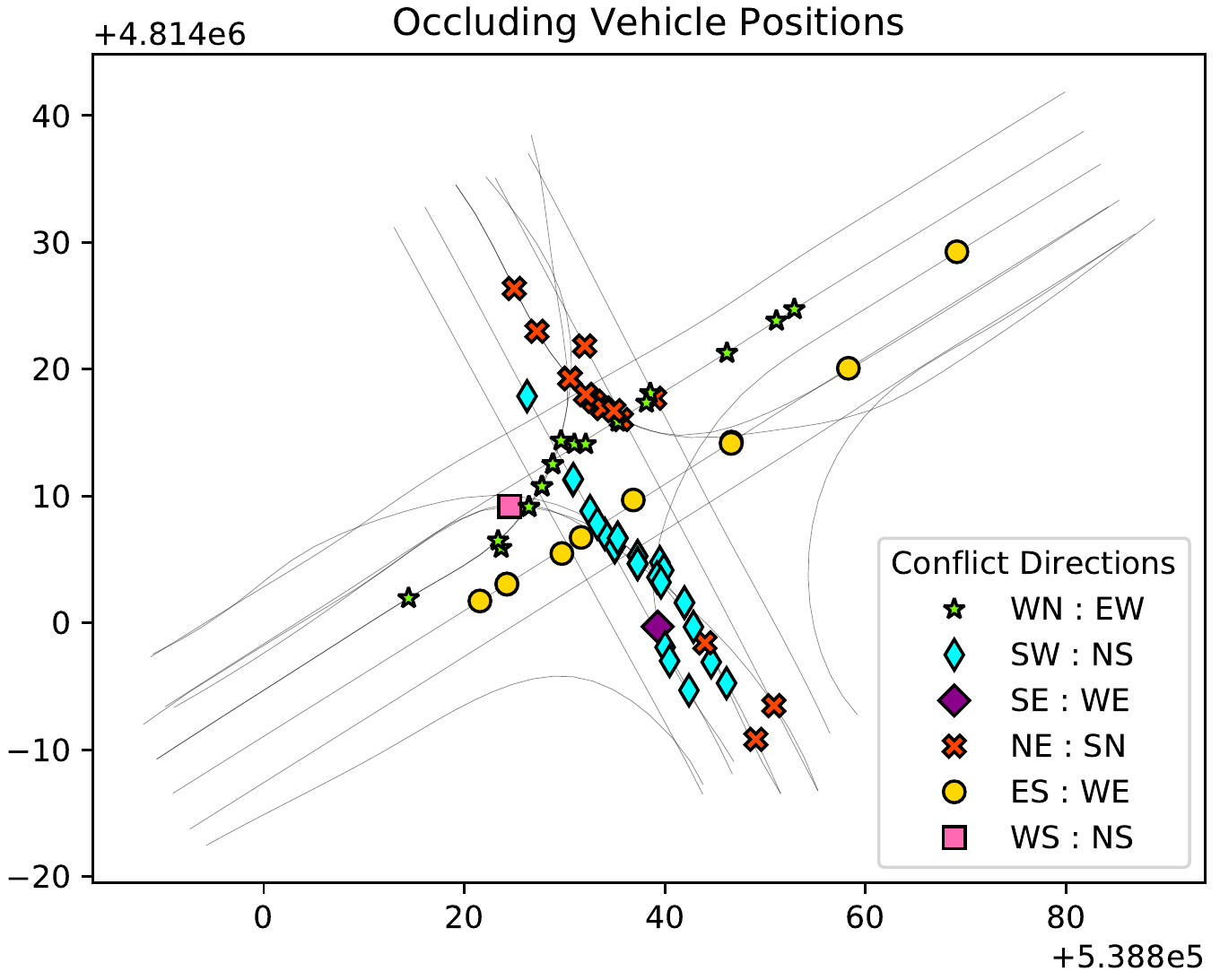}
\label{fig:Occluding Vehicle Positions}} 
\subfloat[]{\includegraphics[width=0.5\linewidth]{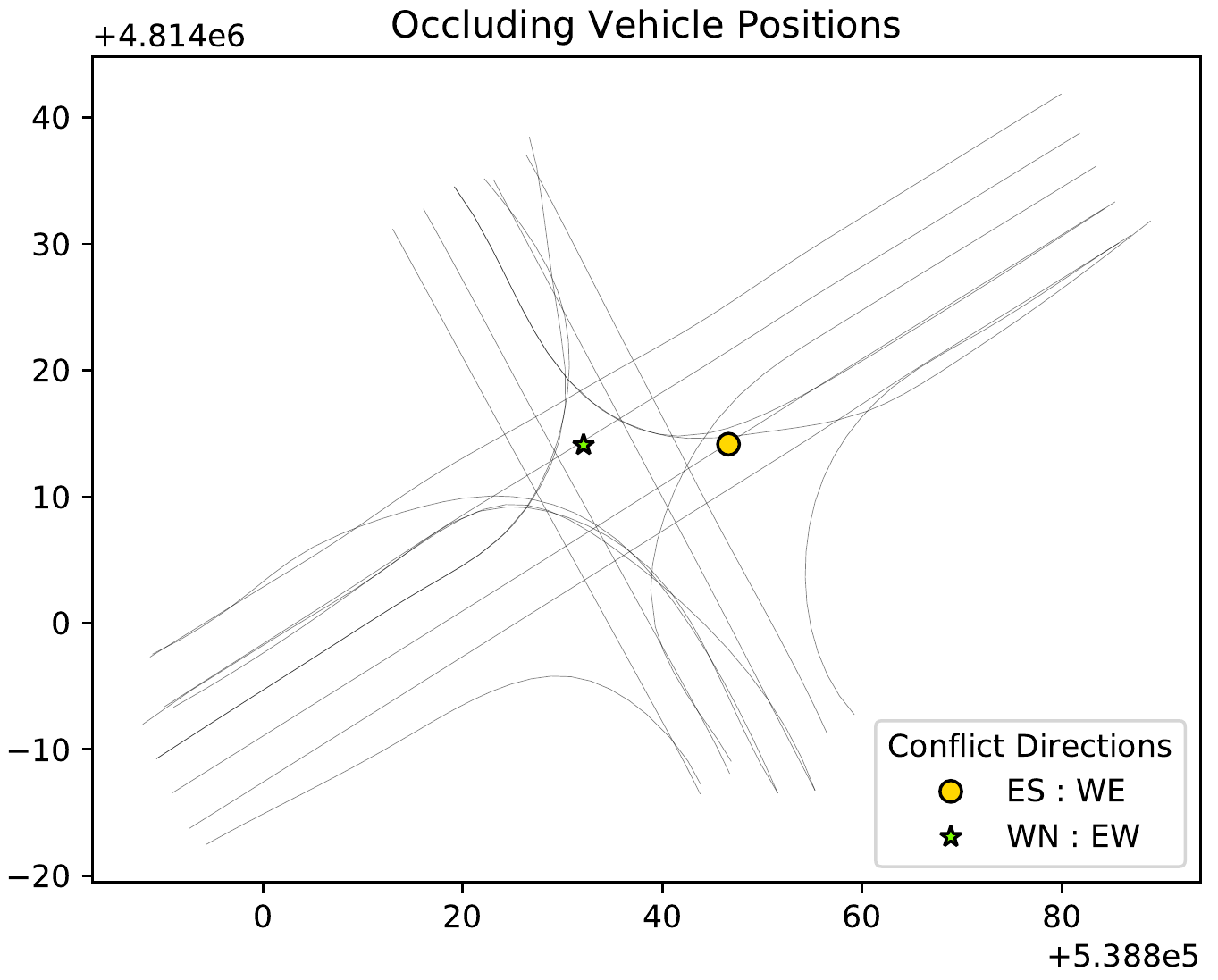}
\label{fig:No SOV Occluding Vehicle Positions}}\newline
\subfloat[]{\includegraphics[width=0.5\linewidth]{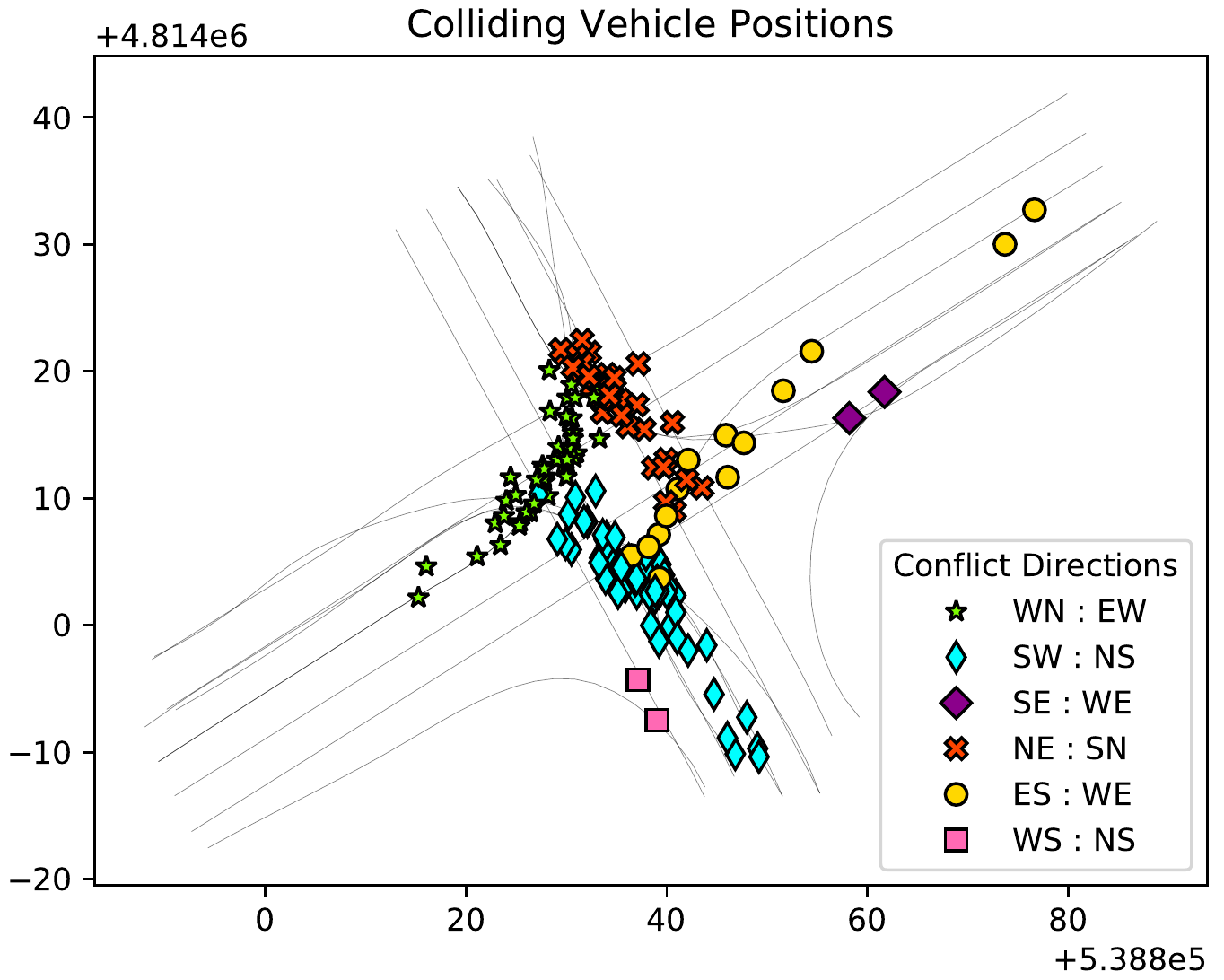}
\label{fig:Colliding Vehicle Positions}}
\subfloat[]{\includegraphics[width=0.5\linewidth]{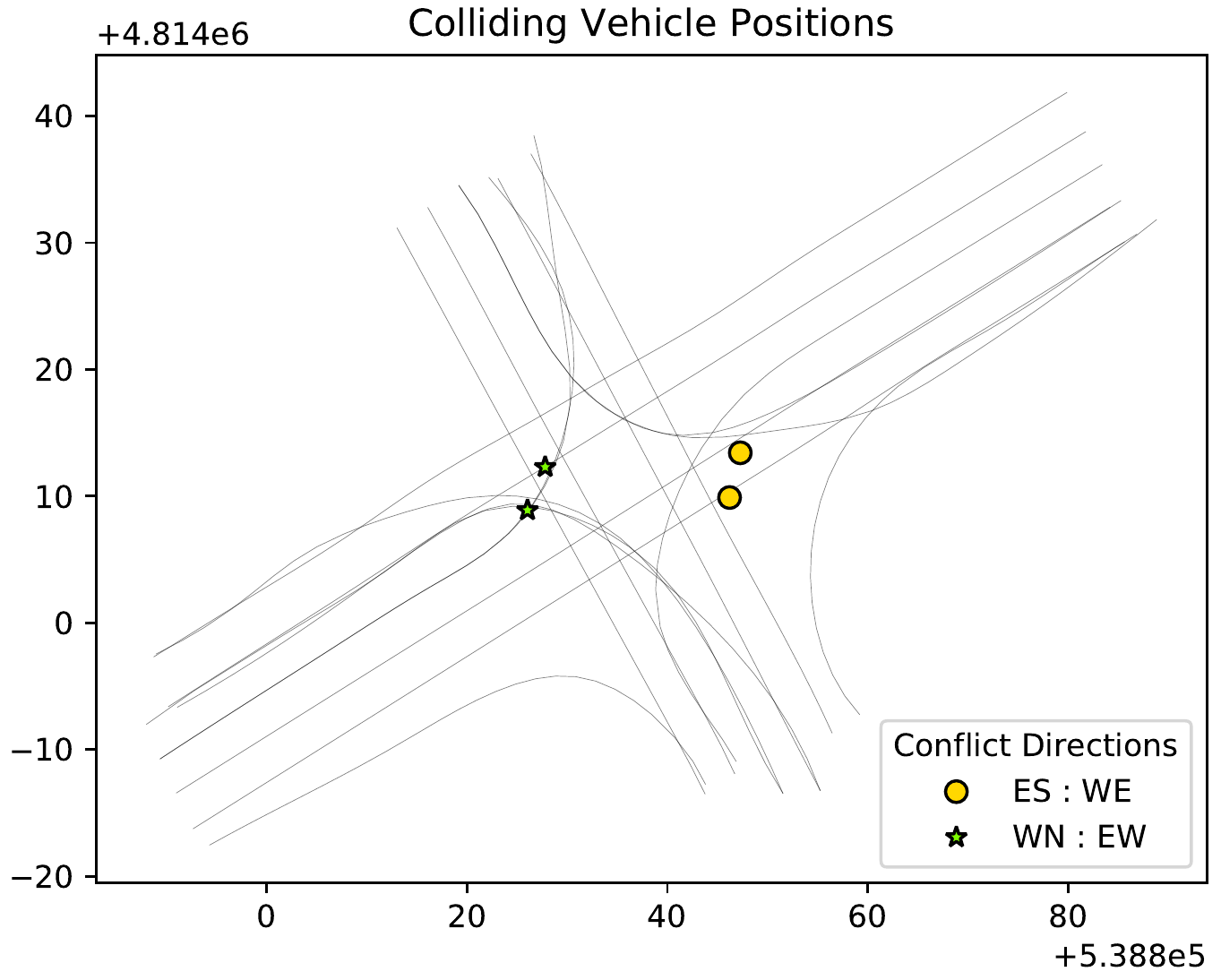}
\label{fig:No SOV Colliding Vehicle Positions}}
\captionof{figure}{Figures (a) and (c) show the occluding vehicle and colliding vehicle positions for the 80 synthetic OCCs. Figures (b) and (d) show the same information but for the 2 naturalistic OCCs. The colliding vehicle positions are plotted at the moment of impact.}
\end{figure}

In addition to faster generation of OCCs, our method also shows a greater coverage of OCC configurations. Figures \ref{fig:Occluding Vehicle Positions} and \ref{fig:No SOV Colliding Vehicle Positions} show the occluding vehicle positions across the 80 OCCs generated using the augmented dataset and the 2 OCCs generated from the non-augmented dataset, respectively. Likewise, figures \ref{fig:Colliding Vehicle Positions} and \ref{fig:No SOV Colliding Vehicle Positions} shows the positions of each vehicle involved in an OCC at the moment of impact for the augmented and non-augmented dataset, respectively. The figures show that our approach achieved a much more even and wider coverage of the dynamic occlusion situations compared to the one observed from naturalistic data.\par
\noindent \textbf{Diversity of generated scenarios}. We were also able to generate a diverse range of OCCs. Amoung the 80 OCCs, we found 53 \textit{front-to-front} collisions, 21 \textit{angle} collisions, 4 \textit{sideswipe} collisions, and 2 \textit{front-to-rear} collisions (refer \cite{national2012mmucc} for details about the these types). Many of these collisions can be attributed to ``tagging on'' behaviour, which is characterized by a vehicle following behind a left-turning occluding vehicle and not noticing that there is an oncoming vehicle. This behavior is typically caused by the follower vehicle making the impatient decision to proceed with the left-turn even though they do not know if there are oncoming vehicles. In fact, 69 out of the 80 OCCs possess this characteristic. This ``tagging on'' is an occlusion-naive behavior, and is observed to be a common cause of crashes due to dynamic occlusion in the real world (\url{https://youtu.be/jDEZ-igoDgw?t=629}). The same behavior was also illustrated earlier in Fig. \ref{fig:Full Scenario 778 4243}, where vehicle 1 performs this ``tagging on'' behavior which results in a collision with vehicle 2.\par
\begin{figure}[tbp]
\centering
\subfloat[]{\includegraphics[trim={0 0 -0.5cm -0.1cm}, width=0.5\linewidth]{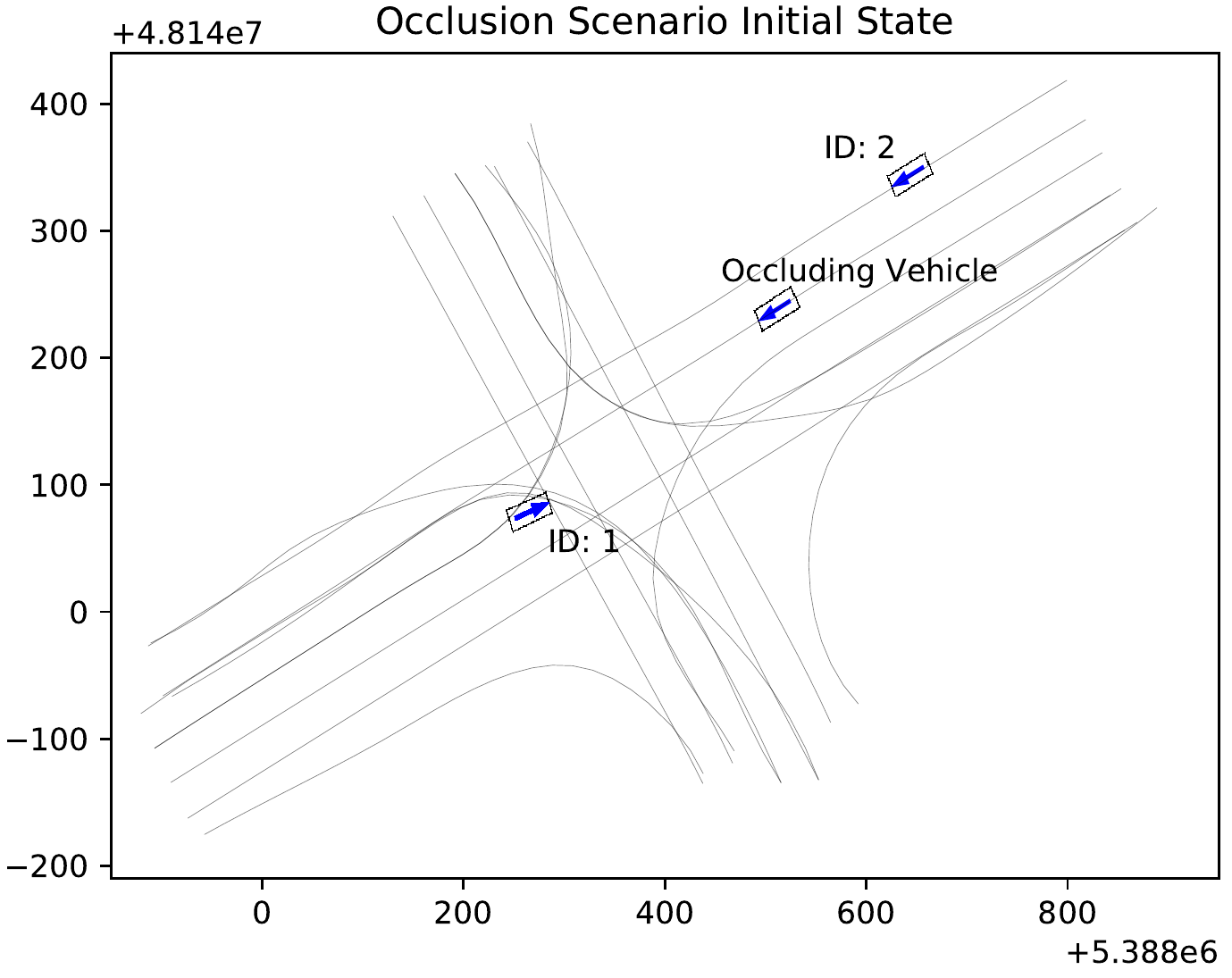}
\label{fig:Occlusion Scenario Initial State}}
\subfloat[]{\includegraphics[trim={0 0 -0.5cm -0.1cm}, width=0.5\linewidth]{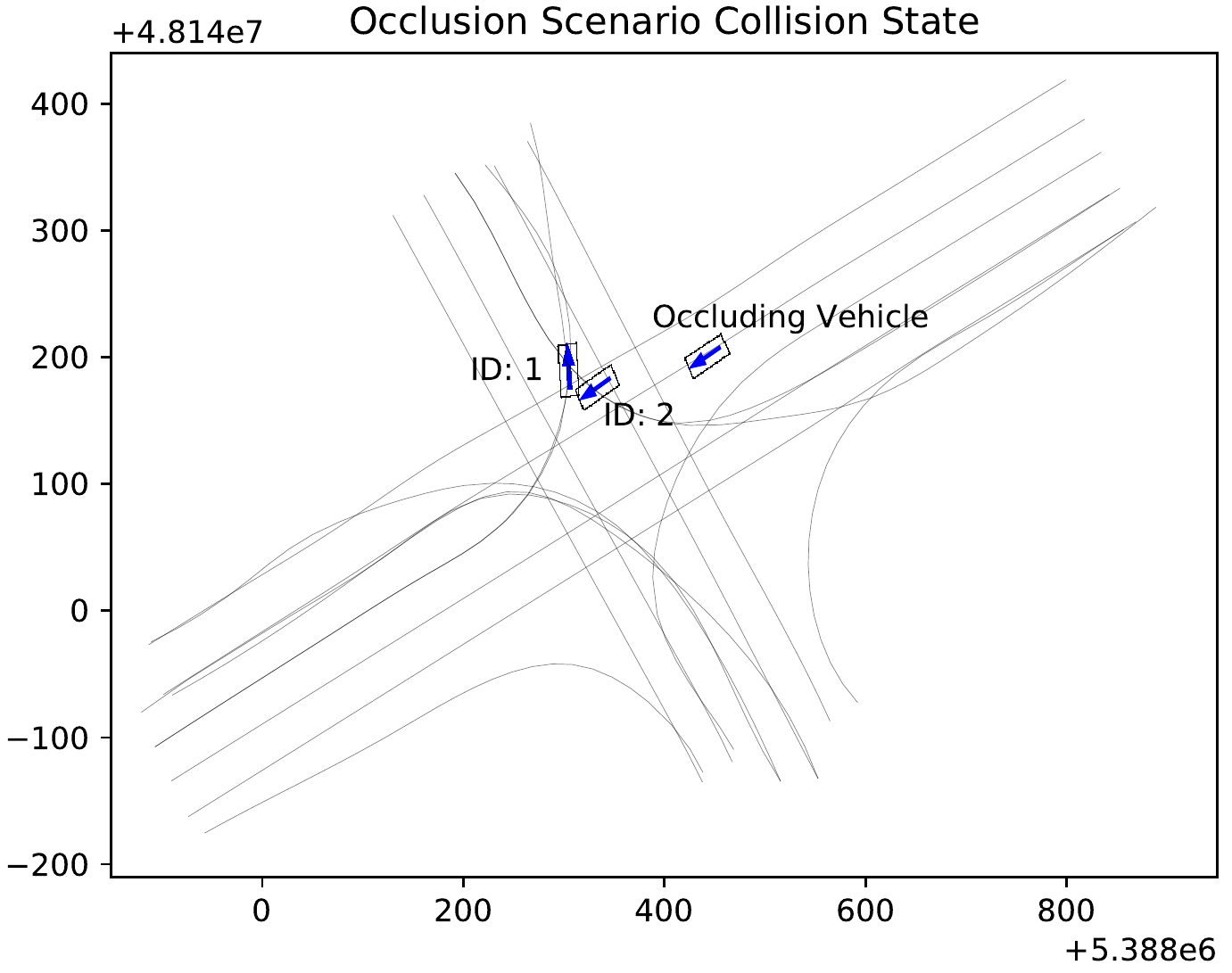}
\label{fig:Occlusion Scenario Collision State}}
\captionof{figure}{(a) 1 begins making a left-turn while 2 proceeds through the intersection. Both vehicles are occluded from each other. (b) 1 collides with 2.}
\end{figure}

11 out of the 80 OCCs do not possess this ``tagging on'' behavior. However, in the case of LTAP scenarios, the collision is also caused by similar occlusion-naive behaviour of the left-turning vehicle, proceeding without having vision of potential oncoming vehicles. Figure \ref{fig:Occlusion Scenario Initial State} is an example of this situation. Here, vehicle 1 executes a left-turn without having vision of vehicle 2. The result is an ``angle'' collision between both vehicles. This type of occlusion scenario configuration is also commonly seen in real world situations (\url{https://youtu.be/tDN-mwNSJc8?t=43}, \url{https://youtu.be/Qk7ejm8MlEc?t=198}).

Analyzing the locations of the occlusions, Figure \ref{fig:Occluding Vehicle Positions} shows that occluding vehicles tend to be positioned close to the center of the intersection (and along their respective straight-through path for LTAP scenarios). However, for RT scenarios, our results suggest that the occluding vehicle must be positioned at a particular location---at the end of the SW left-turn lane for (WS,NS) OCCs or at the end of the ES left-turn lane for (SE,WE) OCCs. The large clusters in Figure \ref{fig:Colliding Vehicle Positions} show that OCCs tend to occur as one of the colliding vehicles is in the process of completing its left-turn and the other colliding vehicle is crossing straight through the intersection. This demonstrates that our proposed method not only has high diversity in the type of risky situations it can identify, but can also generate common situations observed in daily traffic in an automated manner.\par
\begin{figure}[htbp]
\centering
\subfloat[]{\includegraphics[width=0.5\linewidth]{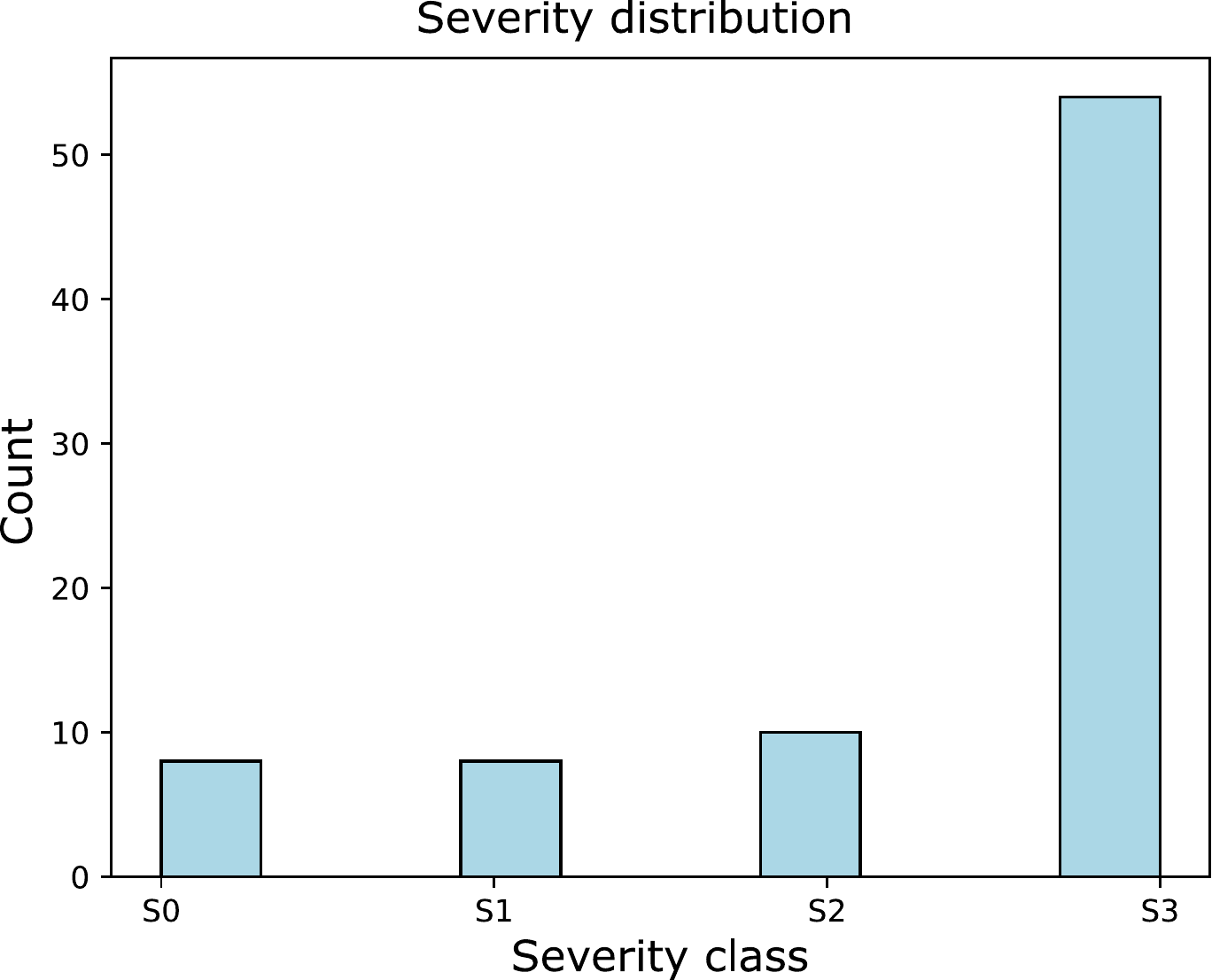}
  \label{fig:Severity Distribution}}
\subfloat[]{\includegraphics[width=0.5\linewidth]{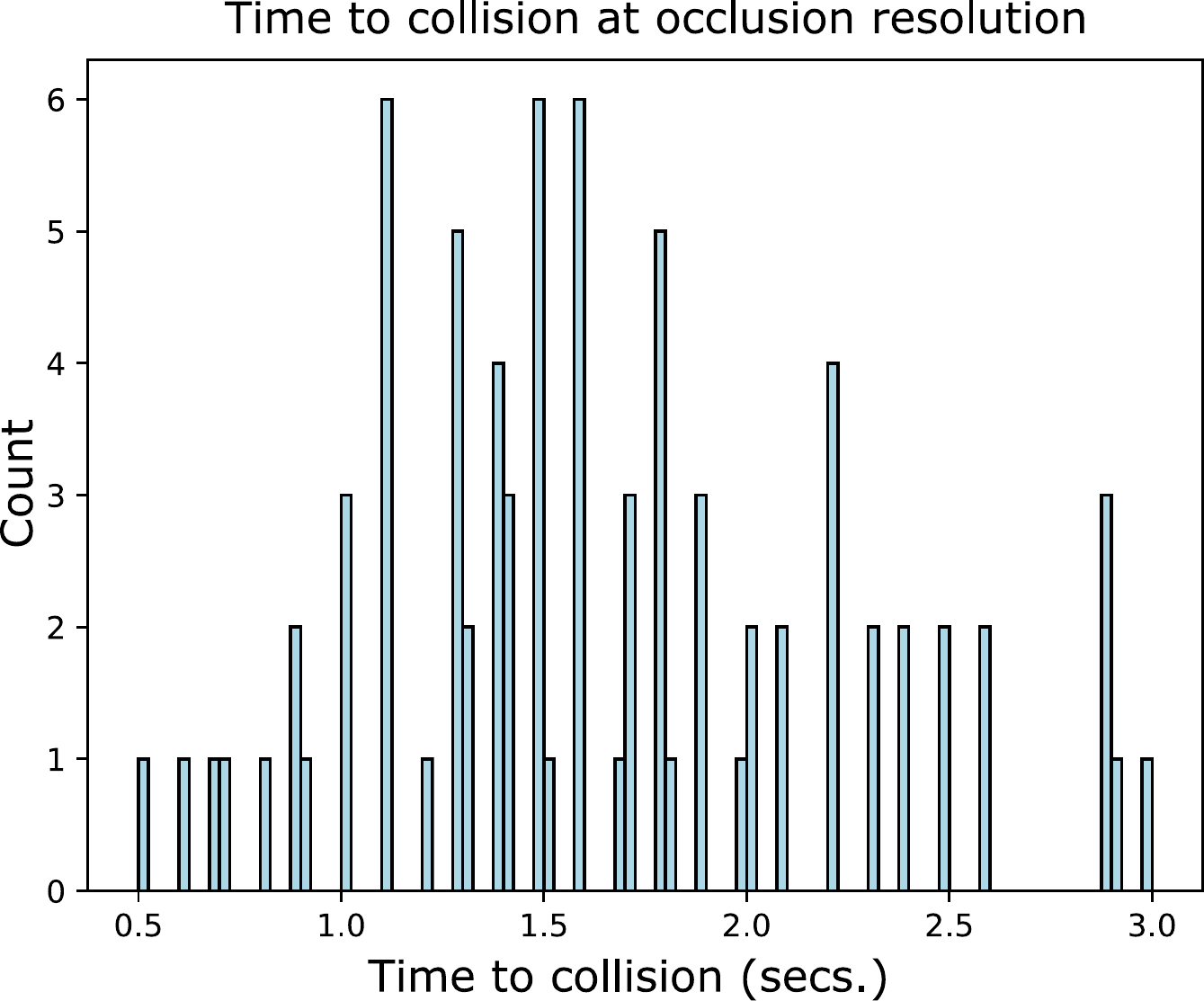}
  \label{fig:Time to Collision After Occlusion}}
\captionof{figure}{(a) Severity class distribution. (b) The distribution of time duration from occlusion resolution to collision.}
\end{figure}
\noindent \textbf{Severity analysis.} Our results also allow for an analysis on how \textit{severe} OCCs tend to be. Figure \ref{fig:Severity Distribution} shows the distribution of the 4 severity classes across the 80 synthetic OCCs. We calculate severity of an OCC by extracting the relative velocity between the colliding vehicles and mapping the value based on the ranges from injury models in \cite{krampe2020injury, sae2018considerations}. The severity class mapping is as follows: S0:[0,5.3], S1:(5.3,7.7], S2:(7.8,10.3], S3:$\geqslant$10.3, all in $\text{ms}^{-1}$. We use relative velocity as a worst-case assumption instead of $\Delta v$ (where $\Delta v$ for a light vehicle approaches the relative velocity between colliding vehicles when the other vehicle in the collision is much heavier, such as a truck). That the severity distribution is skewed towards S3 is not surprising. Out of the 80 OCCs, 78 were LTAP scenarios, which most often lead to front-to-front collisions, resulting in the highest relative impact speeds.
In addition to the high likelihood of a severe collision, OCCs are dangerous because they allow for little time for either driver to respond to the situation. Figure \ref{fig:Time to Collision After Occlusion} shows the distribution of durations from the moment both colliding vehicles are no longer occluded to the moment of impact, across the 80 synthetic OCCs. The distribution ranges from 0.5s to 3.0s with a mean value of 1.65s. The range of driver response time is typically between 0.8s to 2.5s with a mean response time between 1.3s to 1.5s \cite{drozdziel2020drivers, broen1996braking, lerner1993brake}. Based on the average response times then, drivers only have between 0.15s and 0.35s to decelerate or perform an evasive maneuver before a collision. Therefore, proactively identifying potential occlusion situations \textit{before} they occur is critical for ensuring driver safety. A supplementary video containing examples of both synthetic and naturalistic OCCs can be found here: \url{https://youtu.be/-crio3rA_IU}.\par
\section{RELATED WORK}
The literature on occlusion-aware planning can be divided into two categories: reachable set analysis and probabilistic methods. Reachable set analysis \cite{orzechowski2018tackling, koschi2020set, hubmann2019pomdp} provides a method to generate provably-safe trajectories by over-approximating the occupancy states of potential occluded vehicles. A challenge with set-based approaches is to tune the implementation such that the vehicle not only produces provably-safe trajectories but also does not behave too conservatively so as to disrupt the flow of traffic.\par 
Probabilistic models, such as  partially observable Markov decision processes \cite{bouton2018scalable, lin2019decision, hubmann2019pomdp}, provide a method to both handle uncertainty and perform optimal decision-making. Particle filters \cite{yu2019occlusion, narksri2021deadlock}, use Monte Carlo sampling to approximate future positions of potential occluded vehicles, where large clusters of particles indicate a high likelihood of future occupancy. McGill et al. \cite{mcgill2019probabilistic}, propose a probabilistic risk assessment tool which, in addition to incorporating cross traffic, sensor errors and driver attentiveness in its risk calculation, also uses a dynamic Bayesian network to reason about the occupancy of road segments.  Occlusion-aware deep reinforcement learning (DRL) \cite{isele2018navigating, kamran2020risk} has been used to learn safe policies for navigating unsignalized intersections. However, both these works have only been applied to situations with static occlusion and it is difficult for DRL to adapt to unseen scenarios. Unlike reachable set analysis, probabilistic methods do not allow for safety guarantees.\par
In the context of safety validation of AVs, there are several black-box approaches to the problem \cite{o2018scalable,corso2020survey,zhao2016accelerated}. In contrast, our work addresses the problem of safety validation from a white-box perspective, which has received relatively less focus in this domain. Although a very recent work addresses the problem of planning \cite{zhang2021safe}, to our knowledge, there are no existing methodologies for the problem of validation of strategic planners.\par
\section {CONCLUSION}
In this work we presented a novel safety validation framework for strategic planners in AV. We showed how the theory of hypergames can be used to develop a novel multi-agent measure of situational risk associated with dynamic occlusion scenarios. Based on that measure, we developed an accelerated approach of safety validation by augmenting naturalistic datasets with realistic dynamic occlusion scenarios, and assessing the safety of a strategic planner. We showed that the validation method can achieve $10^{4}$ gain in generation of dynamic occlusion scenarios, 4000\% gain in generation of collision scenarios, as well as diversity and alignment with common crash situations. Ultimately, we foresee our proposed method fitting into a larger safety validation pipeline \cite{koren2021approximate}, where, first, failure scenarios are found in low-fidelity in an accelerated manner, followed by a high-fidelity examination of these failure scenarios. We hope that this work can be a stepping stone for performing safety validation of autonomous vehicles under situations with high levels of dynamic occlusion.
\bibliographystyle{IEEEtran}
\bibliography{references}
\end{document}